\documentclass[conference]{IEEEtran}
\IEEEoverridecommandlockouts
\usepackage{cite}
\usepackage{amsmath,amssymb,amsfonts}
\usepackage{graphicx}
\usepackage{textcomp}
\usepackage{xcolor}
\newtheorem{theorem}{Theorem}
\usepackage{booktabs}

\usepackage{multirow,makecell,array}

\usepackage{algorithm}
\usepackage{algpseudocode}
\usepackage{mathtools}
\usepackage{float}
\usepackage{csquotes}
\usepackage{colortbl}
\usepackage{afterpage}
\makeatletter
\let\c@lofdepth\relax
\let\c@lotdepth\relax
\makeatother
\usepackage{subfigure}

\def\BibTeX{{\rm B\kern-.05em{\sc i\kern-.025em b}\kern-.08em
    T\kern-.1667em\lower.7ex\hbox{E}\kern-.125emX}}
\begin{document}

\title{Shuffled Differentially Private Federated Learning for Time Series Data Analytics}

\author{
\IEEEauthorblockN{Chenxi Huang}
\IEEEauthorblockA{\textit{Institute for Infocomm Research}\\
\textit{Agency for Science,}\\
\textit{Technology and Research}\\
138632, Singapore\\
chenxihuang@u.nus.edu}
\and
\IEEEauthorblockN{Chaoyang Jiang}
\IEEEauthorblockA{\textit{School of Mechanical Engineering}\\
\textit{Beijing Institute of Technology}\\
Beijing, 100081, China\\
cjiang@bit.edu.cn}
\and
\IEEEauthorblockN{Zhenghua Chen*}
\IEEEauthorblockA{\textit{Institute for Infocomm Research}\\
\textit{Agency for Science,}\\
\textit{Technology and Research}\\
138632, Singapore\\
chen0832@e.ntu.edu.sg}
}

\maketitle

\begin{abstract}

Trustworthy federated learning aims to achieve optimal performance while ensuring clients' privacy. Existing privacy-preserving federated learning approaches are mostly tailored for image data, lacking applications for time series data, which have many important applications, like machine health monitoring, human activity recognition, etc. Furthermore, protective noising on a time series data analytics model can significantly interfere with temporal-dependent learning, leading to a greater decline in accuracy. To address these issues, we develop a privacy-preserving federated learning algorithm for time series data. Specifically, we employ local differential privacy to extend the privacy protection trust boundary to the clients. We also incorporate shuffle techniques to achieve a privacy amplification, mitigating the accuracy decline caused by leveraging local differential privacy. Extensive experiments were conducted on five time series datasets. The evaluation results reveal that our algorithm experienced minimal accuracy loss compared to non-private federated learning in both small and large client scenarios. Under the same level of privacy protection, our algorithm demonstrated improved accuracy compared to the centralized differentially private federated learning in both scenarios.
\end{abstract}


\section{Introduction}
Federated Learning (FL) ensures privacy by guaranteeing local data storage, yet the risk of privacy breach remains, as attackers can extract sensitive information through reverse analysis during parameter sharing\cite{bb2}\cite{bb3}. Various privacy protection techniques exist, such as Secure Multi-party Computation (SMC) and Homomorphic Encryption (HE), cryptographic methods, and data perturbation techniques like Differential Privacy (DP). HE offers lossless encryption, but it carries considerable computational overhead\cite{bb4}\cite{bb6}. Conversely, DP is a lightweight solution that offers quantifiable and context-free privacy protection for machine learning, thus gaining widespread research interest \cite{bb3}\cite{bb8}. However, the privacy protection offered by DP comes at the cost of utility loss due to noise injection. Thus, it's imperative to develop DP mechanisms that ensure enhanced privacy protection while minimizing noise injection.

Currently, most privacy-preserving FL frameworks cater to text and image data, with limited applicability to time series (TS) data. TS data models or patterns must accommodate the inherent temporal dependencies in the data. Noise injection into the gradients of these models can disrupt these dependencies, causing a more significant interference in the learning process and resulting in a higher accuracy drop compared to conventional learning models. Among the available techniques, those employing HE technology have high computational and communication costs, while those using DP technology fare better \cite{my1}. However, existing DP technology offers inadequate protection for TS data. Some methods involve decomposing TS data and applying DP technology to select components, which may result in valid information loss \cite{my3}. Moreover, these techniques do not consider FL characteristics and cannot handle the requirements for local model updates and attacks originating from multiple sources \cite{my4}.

In response to these challenges, we present a novel privacy-preserving FL algorithm for TS data, based on Local Differential Privacy(LDP). In comparison to DP, LDP provides enhanced protection against semi-trusted servers, preventing privacy leakage from both servers and malicious clients. We also introduce shuffle mechanisms that satisfy LDP while enhancing utility, resulting in privacy amplification \cite{bb17}\cite{bb18}. Additionally, we incorporate the model privacy coefficient, which can be independently configured by each client locally. This adjustment allows for the distribution of the privacy budget between the feature extractor and classifier during local training, thus catering to the unique nature of different clients’ data and their privacy protection requirements.

The experimental results reveal that our algorithm achieved minimal accuracy loss, i.e., 0.9\% for 100 clients and 2.8\% for 1000 clients, compared to non-private federated learning. It also improved accuracy by 7.2\% for 100 clients and 5.9\% for 1000 clients under the same privacy level, compared to centralized DP-based federated learning.

\begin{figure*}[htbp]
  \centering
  \includegraphics[width=0.8\textwidth]{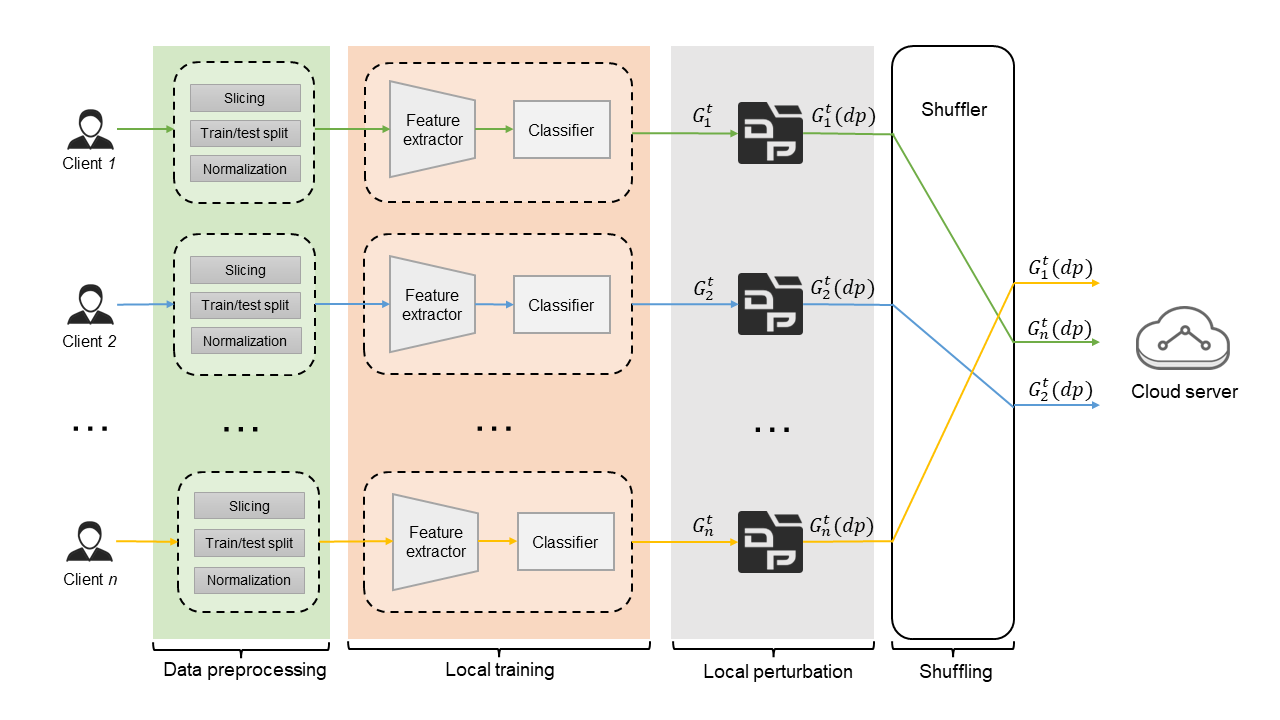}
  \caption{DP-TimeFL Framework}
  \label{f1}
\end{figure*}

Our main contributions include:
\begin{itemize}
\item We propose a novel privacy-preserving FL framework DP-TimeFL for time series based on LDP, ensuring robust privacy safeguards.
\item By implementing model shuffling, we achieve privacy protection amplification while enhancing the utility of the proposed FL framework.
\item We perform comprehensive experiments on five TS datasets to demonstrate our security and accuracy.
\end{itemize}

\section{Related works}\label{sec2}

\subsection{Differentially Private Federated Learning}

DP, a mathematical framework defining privacy properties, is widely used for addressing privacy concerns. Geyer et al. \cite{bb32} proposed a DP-SGD FL framework, offering varying privacy protection levels, while Wei et al. \cite{bb33} introduced NbAFL, meeting global DP requirements by adjusting Gaussian noise variance.

CDP methods assume trusted servers for model aggregation, which may not be reliable in practice \cite{bb34}. LDP addresses FL scenarios with untrusted servers, requiring no trusted third party and realizing privacy protection before data transmission. In LDP, clients store local data and interact individually with untrusted servers \cite{bb21}, keeping local model parameters secret. However, DP mechanisms inherently degrade model performance and utility, with LDP potentially decreasing data availability even more than CDP.

Recently, privacy amplification in LDP-FL has attracted attention \cite{bb36}\cite{bb38}. The shuffling technique weakens the attacker's model, applying LDP to achieve accuracy close to CDP \cite{bb40}\cite{bb41}. By assuming user anonymity, shuffling reduces noise in local models and amplifies privacy.

\subsection{Differentially Private Time Series}

Most private FL frameworks focus on text and image data, with limited works on TS data. A TS feature extraction system was proposed for federated learning using computationally expensive HE \cite{my1}. On the other hand, STL-DP \cite{my3} and PASTE \cite{my4} employ DP for TS privacy. PASTE, designed for data mining, perturbs discrete Fourier transforms of query answers due to TS's temporal correlation. However, it does not suit federated learning's local model updates and cannot defend against untrusted central servers. STL-DP decomposes data using seasonal and trend decomposition with Loess, applying the Fourier perturbation algorithm only to core TS components. This may lead to information loss in the original TS data, and the approach does not consider federated scenarios.






\section{Proposed Framework}\label{sec4}

\subsection{Overview}

The high-level architecture of our DP-TimeFL is shown in Fig. \ref{f1}. We assume that the shuffler and the cloud server are honest but curious, i.e., the shuffler and server perform the shuffling operation and aggregation operation honestly, respectively. However, both may attempt to infer sensitive information from the data uploaded by the client. We consider model inversion attack and membership inference attack in FL scenarios. The $N$ clients independently execute local training processes on their local private data, while the cloud server aggregates local gradients. The goal of the training is to obtain a model with higher model utility than training only on one's local data, without revealing the privacy information of one's local private data. 

As federated learning, the server initializes the global model parameter. Clients perform local training, perturb gradients with Laplace noise, and upload them to the shuffler. The shuffler processes and sends perturbed gradients to the server, which aggregates them to obtain global private gradients. Finally, the server broadcasts the global gradients, allowing clients to update their local weights.

\subsection{Local Perturbation}

The local training model consists of a feature extractor and a classifier. The feature extractor uses a 1D-CNN as the backbone, which is used for TS analytics \cite{ada1}\cite{ada4}.
First, the gradient $g_i^t$ is calculated and clipped. Second, Laplace noise is added to the gradient for perturbation, as shown in \eqref{equ_4}\eqref{equ_5}\eqref{equ_6}.

\begin{equation}
\label{equ_4}
g_{dp}(c_i^t)=\hat{g}(c_i^t)+Lap\left(\frac{(1-k)\Delta f}{\epsilon_t}\right),
\end{equation}

\begin{equation}
\label{equ_5}
g_{dp}(f_i^t)=\hat{g}(f_i^t)+Lap\left(\frac{k\Delta f}{\epsilon_t}\right)
\end{equation}
where $k\in (0, 1)$ is the model privacy coefficient, independently set by each client locally to adjust the privacy budget allocation between the feature extractor and classifier. This adapts to the uniqueness of different clients' data and their privacy protection goals. The larger the $k$, the more privacy budget is allocated to the classifier, the less noise is added to the classifier, and the lower the privacy protection level for the classifier. A larger $k$ value tends to protect the feature extractor rather than the classifier. If $G_i^t$ is a tuple containing the gradients of the feature extractor and classifier, then

\begin{equation}
\label{equ_6}
G_i^t(dp)=\hat{G}_i^t+Lap\left(\frac{\Delta f}{\epsilon_t}\right)
\end{equation}

where $\Delta f$ is the sensitivity and $\epsilon_t$ is the DP parameter of round $t$. The value range of each gradient $G_i^t$ is limited to $(0, 1)$ by min-max normalization, and the sensitivity $\Delta f$ is set to 1 in our method. The DP parameter $\epsilon_t$ is jointly determined by the initial DP parameter $\epsilon_0$ and dynamic adjustment.

\subsection{Privacy Amplification with Shuffling}

In DP-TimeFL, a shuffler is designed according to the existing security shuffle protocol, which treats the shuffler as a black box \cite{bb40}\cite{bb41}. Regarding the specific implementation, the shuffler can be built based on trusted hardware, SMC, or HE with the help of existing secure shuffling protocols according to the model deployment conditions. We randomly replace the encrypted input gradients via the shuffler to achieve LDP, reducing the added noise, realizing privacy amplification, and preventing side-channel linkage attacks in FL. After the client sends the private gradient to the shuffler, for each gradient $G_i^t(dp)$ of the client $C_i$, the shuffler randomly samples a delay $t_i$ from the uniform distribution $U(0,T)$, in which $T$ is a shuffling parameter. When FL is initialized, all clients propose their suggested values $T_i$ to the server. The server takes the median value of all $T_i$ as $T$. For each gradient, the private gradient $G_i^t(dp)$ is uploaded to the server at time $t_i$.

\section{Secure Analysis}\label{sec5}

In contrast to CDP, the shuffled DP approach does not depend on a trusted server and has improved security. Moreover, it fills the $O(\sqrt{n})$ gap in utility between CDP and LDP; that is, shuffled DP can tolerate $O(\sqrt{n})$ times fewer data errors than LDP. In the traditional CDP model, the Laplace mechanism satisfies $\left(\epsilon, 0\right)$-DP, and the error is $\mathrm{O}(\frac{1}{\epsilon})$. In contrast, general LDP meets the error of $\Omega(\frac{1}{\epsilon}\sqrt n)$. The privacy amplification achieved by shuffling can transform the local data that meet $\epsilon_l$-LDP before shuffling into data that meet $\epsilon_c$-DP after shuffling. $\epsilon_l$ corresponds to a larger value, indicating lower privacy; $\epsilon_c$ corresponds to a smaller value, indicating higher privacy. Based on the interactive and non-interactive mechanisms in shuffle DP\cite{bb18}\cite{bb54}, the privacy amplification theorem on the non-general and general interactive mechanisms was proposed in \cite{bb40}\cite{bb41}. It should be noted that the interactive and non-interactive mechanisms here refer to the interaction between users rather than the interaction between iteration rounds. That is, whether the results of a user are affected by the input of other users.
In our setting, DP protection performs in a non-interactive manner between users. Therefore, we adopt the non-interactive shuffle privacy protection mechanism. The general privacy amplification theorem applying a non-general interactive mechanism in  \cite{bb40} is shown as follows.
\begin{theorem}
\label{theorem2}
Given $n$ clients, each with dataset $D_i$, for any $n\in N_+$, $\delta \in [0,\ 1]$,
$\epsilon_l\in \left(0,\frac{1}{2}\ln\frac{n}{\ln\frac{1}{\delta}}\right)$; if the gradient obtained by the local training of the client satisfies $\epsilon_l$-LDP, then the n outputs after shuffling satisfy $(\epsilon_c,\delta)$-DP, where

\begin{equation}
\label{equa_8}
\epsilon_c=\mathrm{O}\left(\left(e^{\epsilon_l}-1\right)\sqrt{\frac{ln{\frac{1}{\delta}}}{n}}\right)
\end{equation}
\end{theorem}

\begin{table}[tbp]
  \centering
  \setlength{\tabcolsep}{8pt}
  \caption{Privacy Amplification Converse Results on a General Non-interactive Mechanism}
  \label{tab1}
  \begin{tabular}{cccccc}
    \toprule
    \multirow{2}{*}{$n$} & \multirow{2}{*}{ }& \multicolumn{4}{c}{$\epsilon_l$}\\
    \cline{2-6}\rule{0pt}{10pt}
    & $\epsilon_c=0.1$ & $\epsilon_c=0.3$ & $\epsilon_c=0.5$ & $\epsilon_c=0.7$ & $\epsilon_c=0.9$\\
    \midrule
   $10^4$  & 1.16 & 2.03 &	2.48 &	2.80 &	3.03\\

    $10^5$  & 2.07 & 3.08 &	3.58 & 3.90 & 4.15\\

    $10^6$ & 3.13 & 4.20	 & 4.71 &	5.04&	5.29\\
    $10^7$ & 4.26 & 5.34	 & 5.85 & 6.19 &	6.44\\
    $10^8$ & 5.40 & 6.49	 & 7.00 &	7.34 &	7.59\\
    \bottomrule
 \end{tabular}
\end{table}

According to Theorem \ref{theorem2}, given $\delta={10}^{-9}$, $n$ trusted clients, and the corresponding value of $\epsilon_c$, we obtain the results of $\epsilon_l$ before amplification according to Equation \ref{equa_8}, as shown in Table \ref{tab1}. Obviously, our strategy satisfies LDP achieving a larger $\epsilon_l$-LDP on the client side and a smaller $\epsilon_c$-DP on the server side through privacy amplification, thereby obtaining a stronger privacy guarantee than CDP. The results also show that privacy amplification increases as the number of participating clients increases.

\section{Experiments}\label{sec6}

\subsection{Datasets}

We selected five datasets from three real-world applications, namely human activity recognition, sleep stage detection, and machine fault diagnosis, for experimental evaluation. Table \ref{tab2} summarizes the details of each dataset, including the total number of samples ($T$), number of classes ($K$), sequence length ($L$), and number of channels ($C$). The description of each dataset is as follows:

\begin{table*}[htbp]
\caption{Experiment settings}
\begin{center}
\begin{tabular}{cccccccc}
\hline
\cline{1-8}\rule{0pt}{10pt}
    Dataset & $N$ & $n$ & $ts$ & $P$ & $B$ & $\epsilon$ & $\delta$\\
\midrule
   UCIHAR  & 100, 1000 & 40 ($N$=100),100 ($N$=1000) & 8400 &	84 & 28 & 8, 10, 100,non-dp & o.1 ($N$=100), 0.001 ($N$=1000)\\

    WISDM  & 100, 1000 & 40 ($N$=100),100 ($N$=1000) & 6000 &	60 & 30 & 8, 10, 100, non-dp & o.1 ($N$=100), 0.001 ($N$=1000)\\

    HHAR & 100, 1000 & 40 ($N$=100),100 ($N$=1000) & 12000 &	120 & 30 & 8, 10, 100, non-dp & o.1 ($N$=100), 0.001 ($N$=1000)\\
    Sleep-EDF & 100 & 34000 &	340 & 34 & 40 & 8, 10, 100, non-dp & o.1\\
    MFD & 100 & 8400 &	84 & 28 & 40 & 8, 10, 100, non-dp & o.1\\
    \bottomrule
\end{tabular}
\label{tab3}
\end{center}
\end{table*}

\begin{figure*}[htbp] \centering
\begin{minipage}[c]{\linewidth}
\subfigure[$N$=100] {
\label{fig1a}
\includegraphics[width=0.48\linewidth,height=0.27\linewidth]{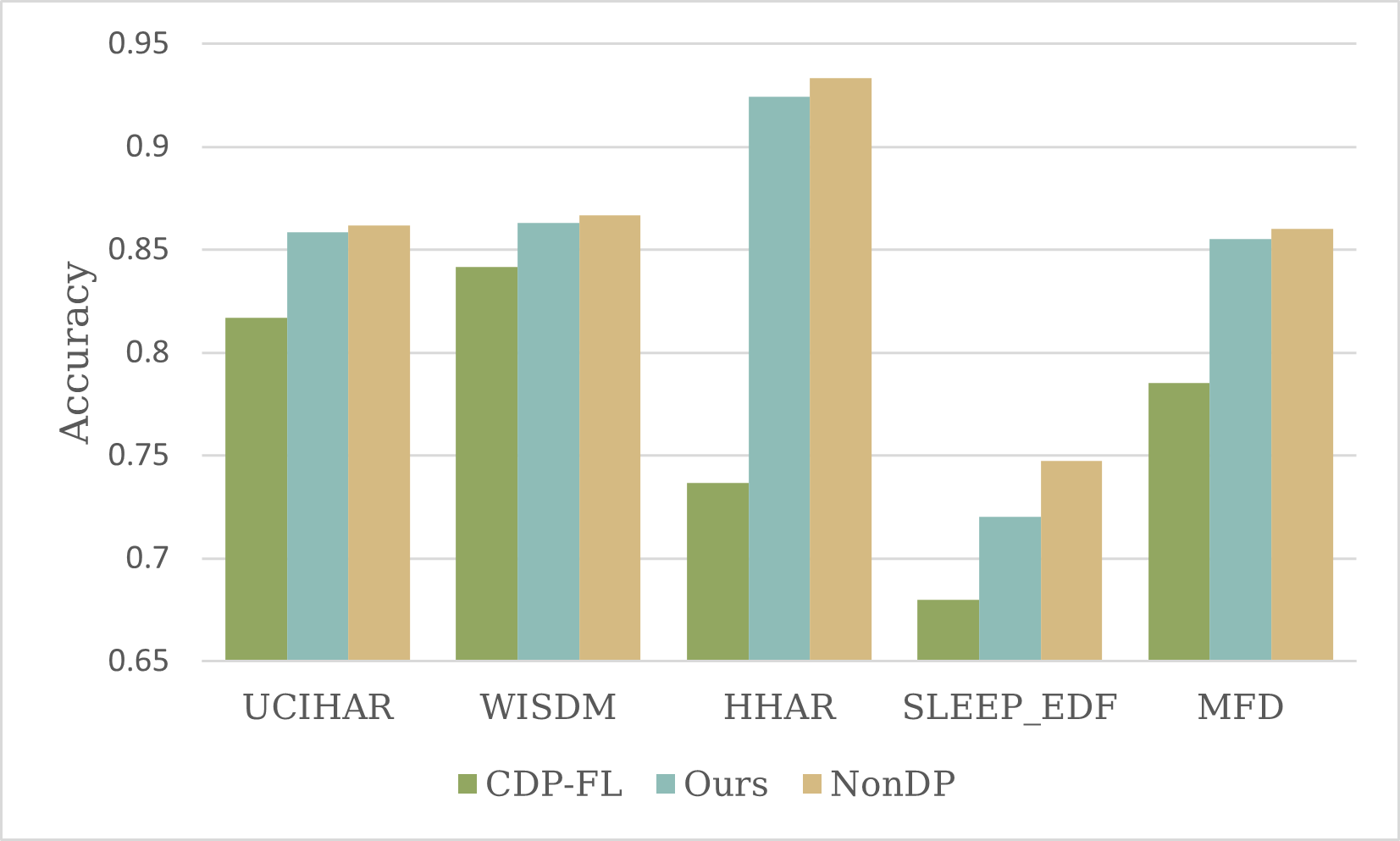}  }
\hspace{0in}
\subfigure[$N$=1000] {
\label{fig1b}
\includegraphics[width=0.48\linewidth]{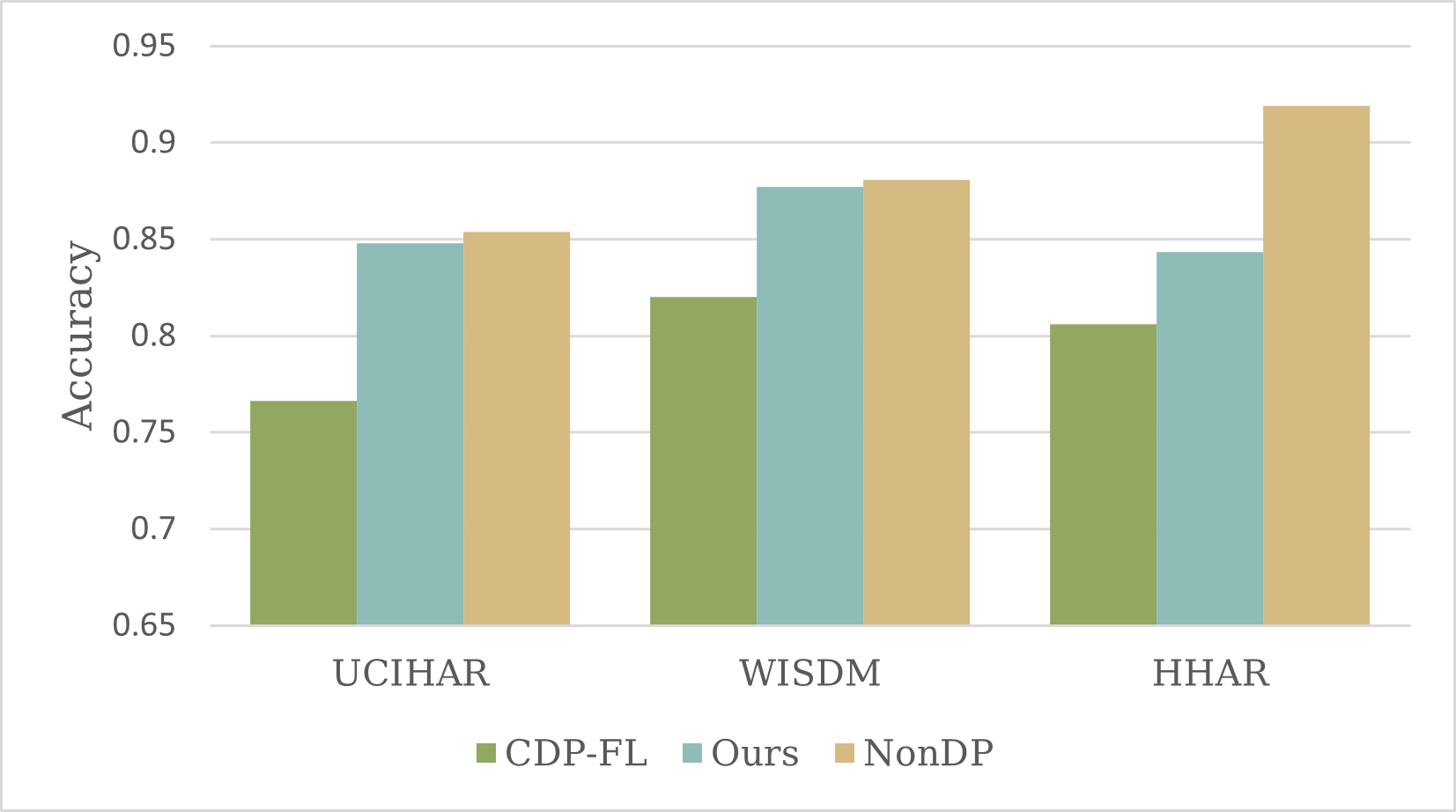}
}
\caption{Global Accuracy on Different Total Clients}
\label{fig1}
\end{minipage}
\end{figure*}

\begin{enumerate}
\item  \textbf{UCIHAR} \cite{ada29} UCIHAR dataset contains data from three sensors, namely accelerometer, gyroscope, and body sensors. Each sensor has three channels. The sensors record data for six activities: walking, walking upstairs, walking downstairs, standing, sitting, and lying down.

 \item  \textbf{WISDM} \cite{ada30} WISDM dataset records the same activities as the UCIHAR with only one three-channel accelerometer for recording, and there is a class imbalance.
 
 \item  \textbf{HHAR} \cite{ada31} The Heterogeneity Human Activity Recognition (HHAR) dataset records sensor readings of activities from heterogeneous smartphones.

 \item   \textbf{Sleep-EDF} \cite{ada32} The Sleep-EDF dataset is used for sleep stage classification tasks, which aims to distinguish electroencephalography (EEG) signals into five stages, i.e., Wake (W), Non-Rapid Eye Movement stages (N1, N2, N3), and Rapid Eye Movement (REM). Following \cite{my5}, we chose the Fpz-Cz channel data for evaluation.
 
 \item  \textbf{MFD} \cite{my7} The Machine Fault Diagnosis (MFD) dataset \cite{my5} has been collected to identify various types of incipient faults using vibration signals. Each sample consists of a single univariate channel with 5120 data points.
 \end{enumerate}

\begin{table}[bp]
\caption{Datasets details}
\begin{center}
\begin{tabular}{ccccc}
\hline
\cline{1-5}\rule{0pt}{10pt}
    Dataset & $T$ & $K$ & $L$ & $C$ \\
\midrule
   UCIHAR  & 10299 & 6 &	128 &	9\\

    WISDM  & 7292 & 6 &	128 & 3\\

    HHAR & 14772 & 6  & 128 &	3\\
    Sleep-EDF & 42308 & 5	 & 3000 & 1\\
    MFD & 10916 & 3	 & 5120 &	1\\
    \bottomrule
\end{tabular}
\label{tab2}
\end{center}
\end{table}

\subsection{Experimental Settings}

We adopt the i.i.d. FL setting in experiments. Table \ref{tab3} summarizes the experimental settings for each dataset, including the number of total clients ($N$), clients participating per round ($n$), the number of training samples ($ts$), data points per client ($P$), batch size ($B$), privacy budget $\epsilon$, and privacy upper limit $\delta$. Each client locally trains 40 epochs. Different total client numbers $N$ correspond to different values of $n$ and $\delta$. Due to the larger size of SLEEP-EDF and MFD datasets, these two datasets use 100 total clients. We perform data slicing, train/test splitting, and normalization in the data preprocessing stage. The ratio of the training set to the test set for each dataset is approximately 0.9:0.1. We use a sliding window of 128 for human activity recognition datasets.

\subsection{Accuracy Evaluation}

\begin{figure*}[htbp] \centering
\begin{minipage}[c]{\linewidth}
\subfigure[UCIHAR] {
\label{fig2a}
\includegraphics[width=0.35\linewidth,height=0.21\linewidth]{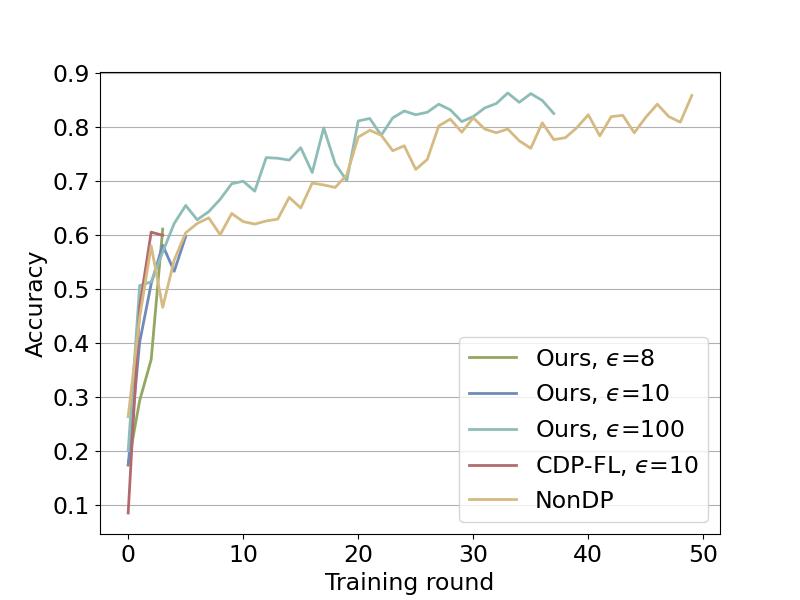}  }
\hspace{-0.4in}
\subfigure[WISDM] {
\label{fig2b}
\includegraphics[width=0.35\linewidth,height=0.21\linewidth]{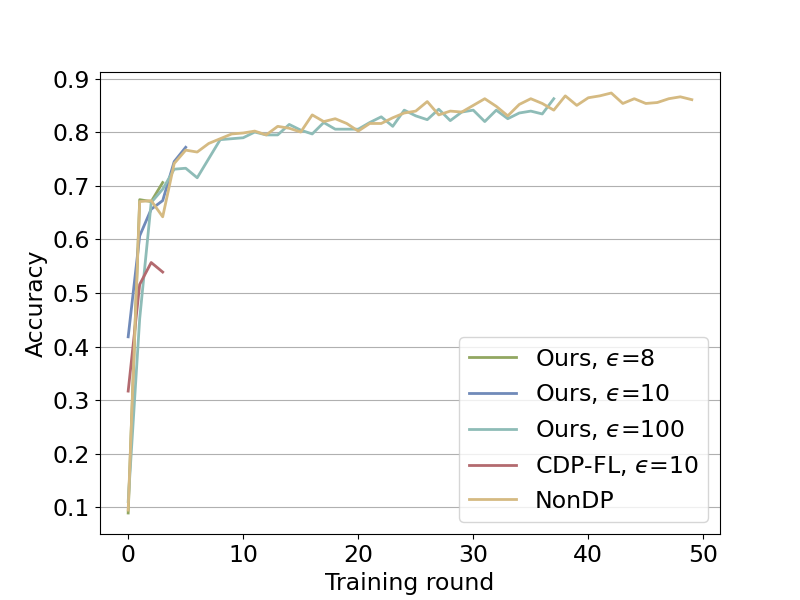}
}
\hspace{-0.4in}
\subfigure[HHAR] {
\label{fig2c}
\includegraphics[width=0.35\linewidth,height=0.21\linewidth]{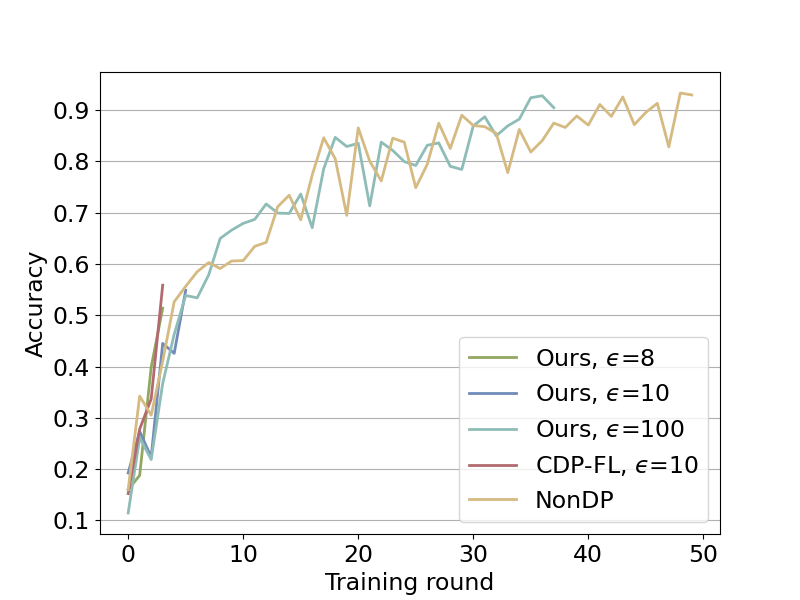}
}
\hspace{-0.4in}
\caption{Global Accuracy on UCIHAR, WISDM, HHAR ($N$=100)}
\label{fig2}
\end{minipage}
\end{figure*}

\begin{figure*}[htbp] \centering
\begin{minipage}[c]{\linewidth}
\subfigure[UCIHAR] {
\label{fig3a}
\includegraphics[width=0.35\linewidth,height=0.21\linewidth]{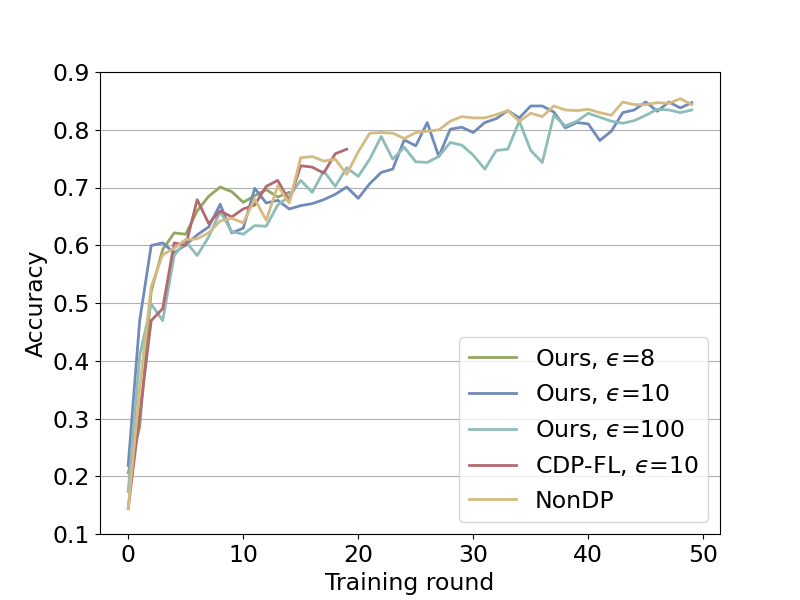}  }
\hspace{-0.4in}
\subfigure[WISDM] {
\label{fig3b}
\includegraphics[width=0.35\linewidth,height=0.21\linewidth]{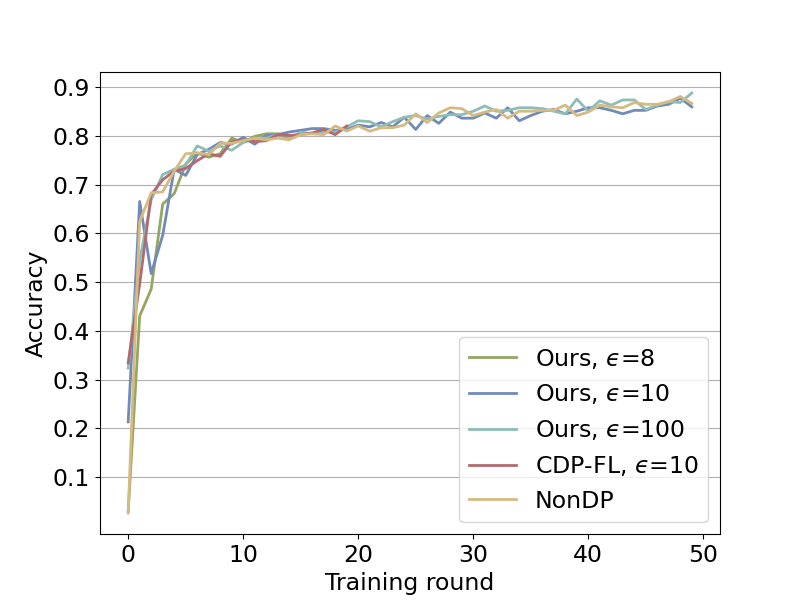}
}
\hspace{-0.4in}
\subfigure[HHAR] {
\label{fig3c}
\includegraphics[width=0.35\linewidth,height=0.21\linewidth]{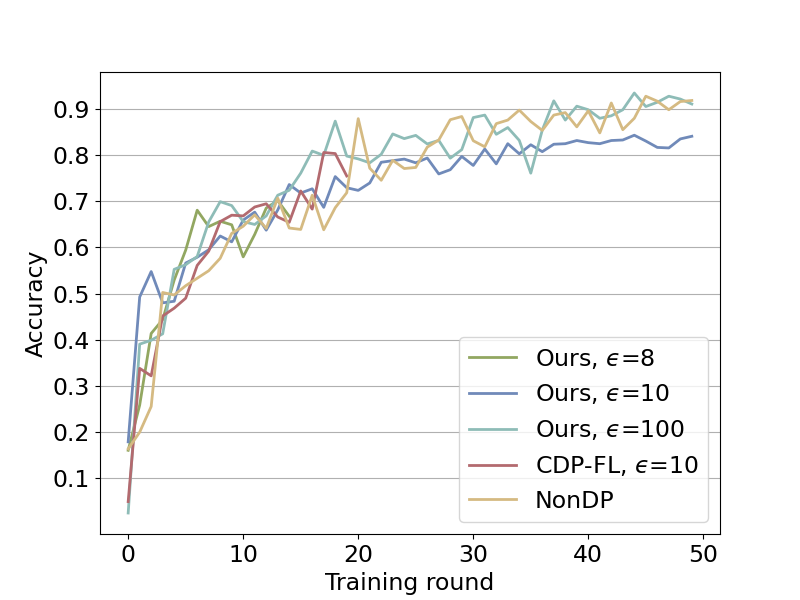}
}
\hspace{-0.4in}
\caption{Global Accuracy on UCIHAR, WISDM, HHAR ($N$=1000)}
\label{fig3}
\end{minipage}
\end{figure*}

\begin{figure*}[htbp] \centering
\begin{minipage}[c]{\linewidth}
\hspace{1in}
\subfigure[SLEEP-EDF] {
\label{fig4a}
\includegraphics[width=0.35\linewidth,height=0.21\linewidth]{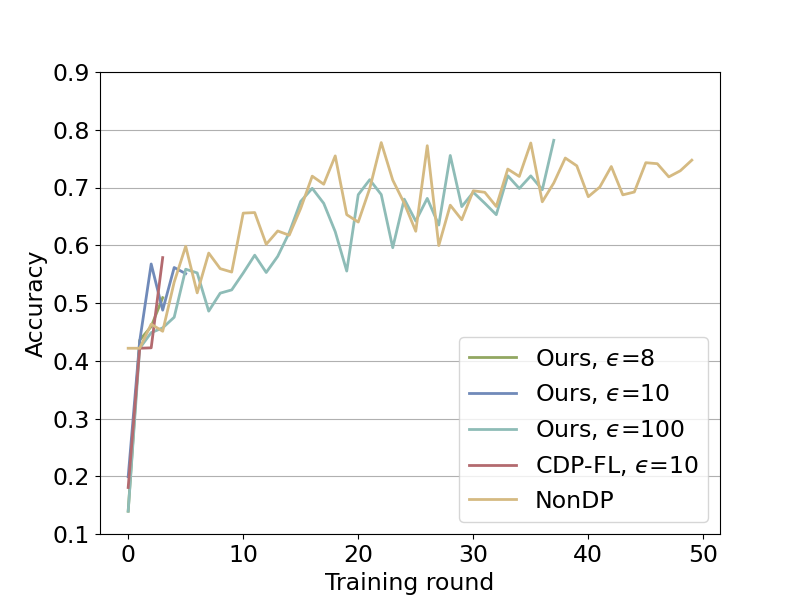}  }
\subfigure[MFD] {
\label{fig4b}
\includegraphics[width=0.35\linewidth,height=0.21\linewidth]{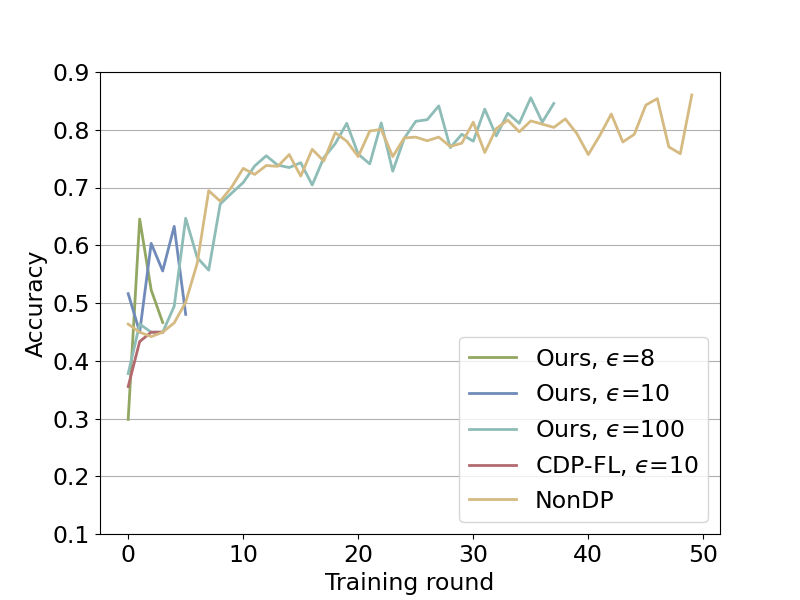}
}
\hspace{-0.4in}
\caption{Global Accuracy on SLEEP-EDF, MFD ($N$=100)}
\label{fig4}
\end{minipage}
\end{figure*}

We conducted experiments comparing DP-TimeFL to FedAvg\cite{bb2} and CDP-FL\cite{bb32}. We initially fixed $\epsilon$ values to 8 and 10 as \cite{my8}. However, these values led to early privacy budget depletion, stopping iteration with low accuracy, especially for fewer clients, as shown in Fig. \ref{fig2}, Fig. \ref{fig3}, and Fig. \ref{fig4}. To address this, we increased the privacy budget to 100.

With a reasonable privacy budget, DP-TimeFL's accuracy on five TS datasets was close to FL without privacy protection, averaging 0.9\% and 2.8\% accuracy loss for 100 and 1000 clients, respectively. As shown in Fig. \ref{fig1}, DP-TimeFL outperformed CDP-based FL by 7.2\% and 5.9\% for 100 and 1000 clients due to privacy amplification through shuffling perturbed gradients. However, the advantage did not become more pronounced with more clients, as the increased privacy amplification was offset by thinner privacy budget distribution, leading to increased noise and decreased accuracy. Thus, as the number of clients increased from 100 to 1000, overall accuracy decreased on most datasets.

\subsection{Privacy Evaluation}

\begin{figure}[tbp]
  \centering
\includegraphics[width=0.5\textwidth,height=0.65\linewidth]{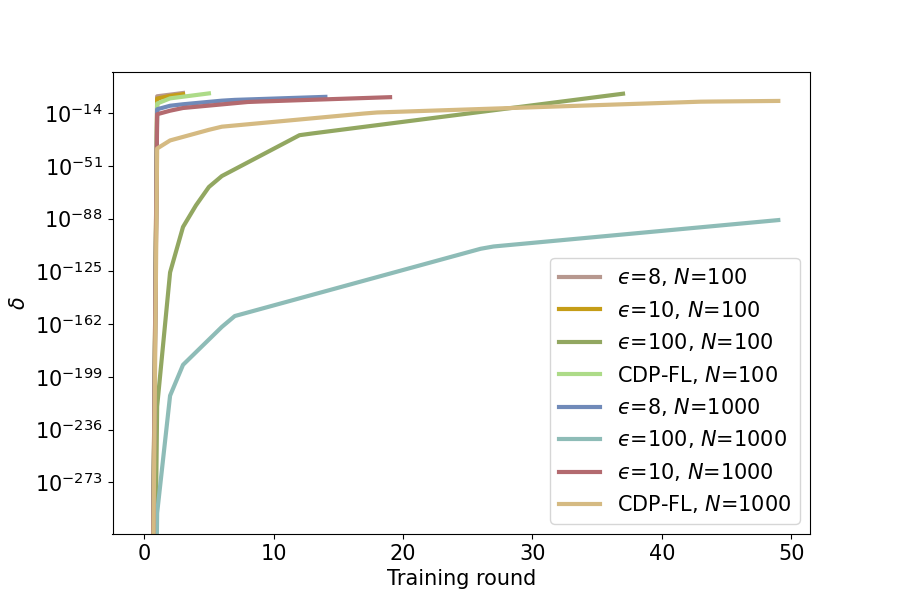}
  \caption{Noise Scale $\delta$ vs. Training Round}
  \label{delta_fig}
\end{figure}

After each round, we calculate the current iteration's $\delta$ using the Gaussian moments accountant and remaining privacy budget. When $\delta$ exceeds the set limit, the iteration stops. As the model converges, gradients become smaller, making noise more impactful, hindering model performance improvement and causing slower $\delta$ convergence in later training stages, as shown in Fig. \ref{delta_fig}. Moreover, we clip gradients to ensure that the L2 norm of each gradient does not exceed a predefined threshold, which helps bound the gradient sensitivity. However, gradient clipping may limit update magnitude in later stages of training, slowing down the $\delta$ convergence. 

With 1000 clients compared to 100 clients, the lower $\delta$ results from each client's smaller contribution to updates, leading to tighter privacy bounds but potentially requiring more communication rounds for desired performance. From another view, more clients improve the slowing convergence speed of $\delta$, benefiting global accuracy by allowing more communication rounds.

\section{Conclusion}\label{sec7}

In this work, we proposed a novel privacy-preserving federated learning algorithm for time series data, which employs LDP. We extended the privacy boundary from the server-side to the client-side to defend against attacks from semi-honest servers. Meanwhile, we introduced a shuffle mechanism to LDP, achieving privacy protection amplification and improving the utility. We conducted experiments on five TS datasets, and the evaluations reveal that our algorithm experienced minimal accuracy loss, with 0.9\% for 100 clients and 2.8\% for 1000 clients, compared to non-private FL. It also improved accuracy by 7.2\% for 100 clients and 5.9\% for 1000 clients under the same privacy level, compared to CDP-based FL.

\section*{Acknowledgment}

This work is supported by the National Natural Science Foundation of China(No.52002026).

\vspace{12pt}


\begin{thebibliography}{00}
\bibitem{bb2} H. B. McMahan, E. Moore, D. Ramage, and B. A. y Arcas, ``Federated learning of deep networks using model averaging," arXiv preprint arXiv:1602.05629, 2016.
\bibitem{bb3} Q. Yang, Y. Liu, Y. Cheng, Y. Kang, T. Chen, and H. Yu, ``Federated learning," Synthesis Lectures on Artificial Intelligence and Machine Learning, vol. 13, no. 3, pp. 1-207, 2019.
\bibitem{bb4} M. Marcantoni, B. Jayawardhana, M. P. Chaher, and K. Bunte, ``Secure formation control via edge computing enabled by fully homomorphic encryption and mixed uniform-logarithmic quantization," IEEE Control. Syst. Lett., vol. 7, pp. 395-400, 2023.
\bibitem{bb6} H. S. A. Fang and Q. Qian, ``Privacy preserving machine learning with homomorphic encryption and federated learning," Future Internet, vol. 13, no. 94, 2021.
\bibitem{bb8} N. Wang, X. Xiao, Y. Yang, J. Zhao, S. C. Hui, H. Shin, J. Shin, and G. Yu, ``Collecting and analyzing multidimensional data with local differential privacy," in 2019 IEEE 35th International Conference on Data Engineering (ICDE), pp. 638-649, 2019.
\bibitem{my1} S. Wang, J. Li, M. Lu, Z. Zheng, Y. Chen, and B. He, ``A System for Time Series Feature Extraction in Federated Learning," in Proceedings of the 31st ACM International Conference on Information \& Knowledge Management (CIKM '22), New York, NY, USA, pp. 5024–5028, 2022.
\bibitem{my3} K. Kim, M. Kim, and S. Woo, ``STL-DP: Differentially Private Time Series Exploring Decomposition and Compression Methods," in Proceedings of the CIKM 2022 Workshops co-located with 31st ACM International Conference on Information and Knowledge Management (CIKM 2022), Atlanta, USA, October, 2022.
\bibitem{my4} V. Rastogi and S. Nath, ``Differentially private aggregation of distributed time-series with transformation and encryption," in Proceedings of the 2010 ACM SIGMOD International Conference on Management of Data, ser. SIGMOD '10, Indianapolis, Indiana, USA, pp. 735-746, 2010.
\bibitem{bb17}  S. P. Kasiviswanathan, H. K. Lee, K. Nissim, S. Raskhodnikova, and A. Smith, ``What can we learn privately?'' SIAM Journal on Computing, vol. 40, no. 3, pp. 793-826, 2011.
\bibitem{bb18} J. C. Duchi, M. I. Jordan, and M. J. Wainwright, ``Local privacy, data processing inequalities, and statistical minimax rates,'' Computer Science, 2013.
\bibitem{bb21} A. M. Girgis, D. Data, S. N. Diggavi, A. T. Suresh, and P. Kairouz, ``On the renyi differential privacy of the shuffle model,'' CoRR abs/2105.05180, 2021.
\bibitem{bb32} R. Geyer, T. Klein, and M. Nabi, ``Differentially private federated learning: A client level perspective,'' ArXiv abs/1712.07557, 2017.
\bibitem{bb33} K. Wei, J. Li, M. Ding, C. Ma, H. H. Yang, F. Farokhi, S. Jin, T. Q. S. Quek, and H. V. Poor, ``Federated learning with differential privacy: Algorithms and performance analysis,'' IEEE Transactions on Information Forensics and Security, vol. 15, pp. 3454-3469, 2020.
\bibitem{bb34} P. C. M. Arachchige, D. Liu, S. A. Çamtepe, S. Nepal, M. Grobler, P. Bertók, and I. Khalil, ``Local differential privacy for federated learning in industrial settings,'' ArXiv abs/2202.06053, 2022.
\bibitem{bb36} Ú. Erlingsson, V. Feldman, I. Mironov, A. Raghunathan, K. Talwar, and A. Thakurta, ``Amplification by shuffling: From local to central differential privacy via anonymity,'' arXiv preprint arXiv:1811.12469, 2018.
\bibitem{bb38} A. M. Girgis, D. Data, S. Diggavi, P. Kairouz, and A. T. Suresh, ``Shuffled model of federated learning: Privacy, communication and accuracy tradeoffs,'' arXiv preprint arXiv:2008.07180, 2020.
\bibitem{bb40} B. Balle, J. Bell, A. Gascón, and K. Nissim, "The privacy blanket of the shuffle model," in Advances in Cryptology – CRYPTO 2019, A. Boldyreva and D. Micciancio, Eds. Cham: Springer, pp. 638–667, 2019.
\bibitem{bb41} Ú. Erlingsson, V. Feldman, I. Mironov, A. Raghunathan, K. Talwar, and A. Thakurta, "Amplification by shuffling: From local to central differential privacy via anonymity," in Proceedings of the 2019 Annual ACM-SIAM Symposium on Discrete Algorithms (SODA), pp. 2468–2479, 2019.
\bibitem{bb54} J. C. Duchi, M. J. Wainwright, and M. I. Jordan, ``Minimax optimal procedures for locally private estimation,'' arXiv preprint arXiv:1604.02390, 2016.
\bibitem{my5} E. Eldele, Z. Chen, C. Liu, et al., ``An Attention-Based Deep Learning Approach for Sleep Stage Classification With Single-Channel EEG,'' IEEE Trans. Neural Syst. Rehabil. Eng., vol. 29, pp. 809-818, 2021.
\bibitem{my7} C. Lessmeier, J. K. Kimotho, D. Zimmer, and W. Sextro, ``Condition monitoring of bearing damage in electromechanical drive systems by using motor current signals of electric motors: A benchmark data set for data-driven classification,'' in PHM Society European Conference, vol. 3, pp. 05-08, 2016.
\bibitem{my8} M. Abadi, A. Chu, I. Goodfellow, H. B. McMahan, I. Mironov, K. Talwar, and L. Zhang, ``Deep Learning with Differential Privacy,'' arXiv preprint arXiv:1607.00133, 2016.
\bibitem{ada1} H. I. Fawaz, G. Forestier, J. Weber, L. Idoumghar, and P. A. Muller, ``Deep learning for time series classification: A review,'' Data Min. Knowl. Discov., vol. 33, no. 4, pp. 917–963, 2019.
\bibitem{ada4} E. Eldele, M. Ragab, Z. Chen, M. Wu, C. K. Kwoh, X. Li, and C. Guan, ``Time-series representation learning via temporal and contextual contrasting,'' in Proc. Thirtieth Int. Joint Conf. Artif. Intell., IJCAI-21, pp. 2352–2359, 2021.
\bibitem{ada29} D. Anguita, A. Ghio, L. Oneto, X. Parra, and J. L. Reyes-Ortiz, ``A public domain dataset for human activity recognition using smartphones,'' in Eur. Symp. Artif. Neural Netw., pp. 437–442, 2013.
\bibitem{ada30} J. R. Kwapisz, G. M. Weiss, and S. A. Moore, ``Activity recognition using cell phone accelerometers,'' SIGKDD Explor., vol. 12, no. 2, pp. 74–82, 2011.
\bibitem{ada31} A. Stisen, H. Blunck, S. Bhattacharya, T. S. Prentow, M. B. Kjærgaard, A. Dey, T. Sonne, and M. M. Jensen, ``Smart devices are different: Assessing and mitigating mobile sensing heterogeneities for activity recognition,'' in Proc. 13th ACM Conf. Embed. Netw. Sens. Syst., pp. 127–140, 2015.
\bibitem{ada32} A. L. Goldberger, L. A. N. Amaral, L. Glass, J. M. Hausdorff, P. C. Ivanov, R. G. Mark, J. E. Mietus, G. B. Moody, C. K. Peng, and H. E. Stanley, ``Physiobank, physiotoolkit, and physionet components of a new research resource for complex physiologic signals,'' Circulation, vol. 101, no. 23, pp. 215–220, 2000.

\end{thebibliography}
\end{document}